\documentclass[acmtog]{acmart} 
\acmSubmissionID{1680}

\def\MethodName{BFS}

\newcommand{\Fig}[1]  {Fig.\ #1}

\newcommand{\Tbl}[1]  {Tab.\ #1}

\newcommand{\Sec}[1]  {Sec.\ #1}

\usepackage{booktabs} 
\usepackage{graphicx}
\usepackage[capitalize]{cleveref}
\usepackage{amsmath}
\usepackage{booktabs}
\usepackage{multirow}
\usepackage{makecell}
\usepackage{duckuments}
\usepackage[ruled]{algorithm2e} 

\SetAlFnt{\small}
\SetAlCapFnt{\small}
\SetAlCapNameFnt{\small}
\SetAlCapHSkip{0pt}

\copyrightyear{2026}
\acmYear{2026}
\setcopyright{cc}
\setcctype{by}
\acmConference[SIGGRAPH Conference Papers '26]{Special Interest Group on Computer Graphics and Interactive Techniques Conference Conference Papers}{July 19--23, 2026}{Los Angeles, CA, USA}
\acmBooktitle{Special Interest Group on Computer Graphics and Interactive Techniques Conference Conference Papers (SIGGRAPH Conference Papers '26), July 19--23, 2026, Los Angeles, CA, USA}
\acmDOI{10.1145/3799902.3811228}
\acmISBN{979-8-4007-2554-8/2026/07}

\begin{document}
\title{BFS: Back-to-Front Layered Image Synthesis via Knowledge Transfer}

\author{Kyoungkook Kang}
\affiliation{%
 \institution{SAMSUNG}
 \country{Republic of Korea}}
\email{kkook.kang@samsung.com}

\author{Gyujin Sim}
\affiliation{%
 \institution{POSTECH}
 \country{Republic of Korea}}
\email{sgj0402@postech.ac.kr}

\author{Sunghyun Cho}
\affiliation{%
 \institution{POSTECH}
 \country{Republic of Korea}}
\email{s.cho@postech.ac.kr}

\begin{abstract}

As generative models expand the possibilities of visual content creation, layered image synthesis has emerged as a promising direction for controllable and creative editing.
However, existing methods struggle to fully realize this potential. Decomposition-based methods often struggle with clean separation, while generation-based methods suffer from difficulty in training data acquisition, reducing quality and scene diversity.
In this paper, we propose \MethodName{}, a novel generation-based framework for layered image synthesis.
Specifically, given a background image and user guidance, \MethodName{} synthesizes a foreground layer that incorporates not only a foreground object but also its associated visual effects, such as shadows and reflections, while seamlessly harmonizing with the background to produce a coherent composite.
To enable diverse and high-quality foreground layer synthesis while overcoming data scarcity, we leverage the comparatively easy-to-learn knowledge of unlayered image synthesis for the foreground synthesis.
To this end, we adopt a dual-branch diffusion framework in which two interconnected branches generate a composite image and a foreground layer, respectively, enabling bidirectional knowledge transfer.
Based on this framework, we propose a two-stage training scheme that utilizes a high-quality unlayered composite image dataset to effectively enhance foreground quality.
Extensive experiments, including a user study, show that \MethodName{} produces high-quality layered images, consistently outperforming prior methods.

\end{abstract}

%
\ccsdesc[500]{Computing methodologies~Image manipulation}

%
%

\begin{teaserfigure}
    \centering
    \includegraphics[width=\linewidth]{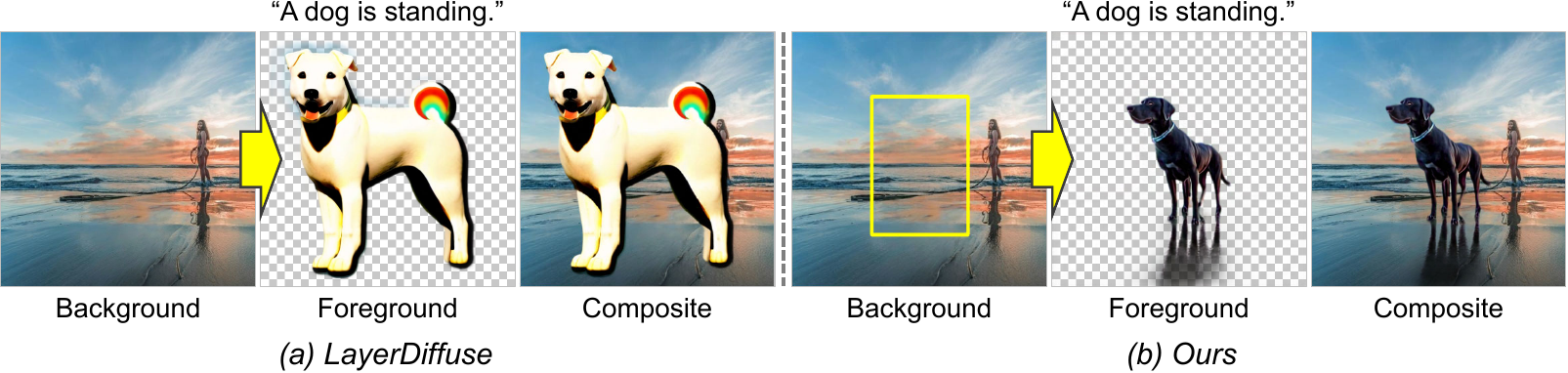}
    \caption{Qualitative comparison of back-to-front layered image synthesis between LayerDiffuse~\citep{layerdiffuse} and our method.
    Given a background image, each method synthesizes a foreground layer and produces a composite by combining it with the background.
    The yellow box denotes the input mask, which is not used by LayerDiffuse.
    LayerDiffuse generates oversized foregrounds inconsistent with the background context, whereas our method produces contextually appropriate foreground layers with proper visual effects, including reflection and shadow.}
    \Description{teaser}
    \label{fig:teaser}
\end{teaserfigure}
\maketitle

\section{Introduction}
\label{sec:introduction}

Layered images, typically organized as a stack of multiple RGBA layers, have long been a standard representation in professional image creation and editing.
Such a representation enables independent manipulation of visual elements without affecting the rest of the image, and this non-destructive property not only streamlines iterative creative workflows but also enables dynamic reuse of visual elements across different compositions.

Recent advances in image generative models~\citep{dalle2,imagine,ldm,sdxl} have opened new opportunities for automatic layered image synthesis. According to their layer construction strategy, existing approaches can be broadly categorized into two paradigms: \emph{decomposition-} and \emph{generation-based} methods.
\emph{Decomposition-based} methods~\citep{layeringdiff,layerdecomp} break down an existing composite image into its constituent foreground and background layers, typically guided by user-provided foreground masks. 
However, these methods often struggle to accurately extract foreground objects together with their associated visual effects, such as shadows and reflections.
Moreover, they are inherently limited to separating existing images and lack the ability to synthesize new layers to construct novel compositions. 

In contrast, \emph{generation-based} methods~\citep{text2layer,layerdiff,layerdiffuse,layerfusion,dreamlayer,art,psdiffusion} directly synthesize layers from random noise and text descriptions.
They produce sharp and accurate alpha masks and allow the straightforward creation of new layered images using text description, which has driven growing research interest.
However, as these approaches require large-scale layered image datasets for training, which are difficult to obtain, the development of a scalable data construction pipeline has emerged as a central challenge. 

For example, Text2Layer~\citep{text2layer}, the MuLAn dataset\\~\citep{mulan}, and DreamLayer~\citep{dreamlayer} propose separation-based data construction pipelines.
Starting from a composite image dataset, they extract foreground layers and then inpaint the residual regions to produce background layers.
However, this process often leaves unnatural artifacts in the background, such as shadows and reflections of absent foreground objects, which in turn causes the trained model to generate incorrect outputs. 
Although DreamLayer~\citep{dreamlayer} filters out low-quality layered images through manual review, such reliance on human labor makes the approach unsuitable for scalable dataset construction.

Other pipelines proposed by \citet{layerdiffuse} and \citet{psdiffusion} start from an RGBA foreground dataset. Specifically, \citet{layerdiffuse} first apply outpainting outside the sampled foreground object cutouts to obtain composite images, then remove and inpaint these cutouts to produce background layers.
However, the RGBA foreground dataset is limited in both diversity and scale, making the resulting composite images less representative of real-world imagery.
Consequently, the generated layered images are confined to a narrow range of object categories and scene variations. 
Meanwhile, \citet{psdiffusion} hired professional designers to select assets from foreground and background datasets and to carefully combine them, but this labor-intensive curation is prohibitively costly.
These limitations underscore the persistent challenges in layered image synthesis, especially in acquiring high-quality data at scale without substantial human involvement.

In this paper, we propose \MethodName{} (Back-to-Front Layered Image Synthesis), a novel generation-based framework that focuses on background-to-foreground (BG2FG) synthesis, i.e., synthesizing a foreground layer conditioned on a given background image.
While alternative strategies exist for layered image synthesis, such as generating all layers simultaneously or synthesizing foregrounds before backgrounds, BG2FG synthesis offers two distinct advantages.
First, it seamlessly extends to multi-layered image synthesis by sequentially adding new layers.
Moreover, it provides a natural design workflow, where users can explore diverse layer compositions by adding objects into a scene one at a time.

To overcome the data scarcity, we leverage relatively easy-to-learn knowledge of unlayered image synthesis to guide the more challenging foreground synthesis.
To this end, \MethodName{} introduces a semi-supervised learning approach based on a dual-branch diffusion architecture. 
Given a background image, our model employs two interconnected branches: one generates a foreground layer that harmonizes with the background, while the other produces the corresponding composite image. This architecture enables bidirectional information exchange between the branches, which facilitates knowledge transfer across modalities, improving both branches.

We train this framework in two stages. In the first stage, we use a synthetic layered image dataset, albeit of lower quality, to teach the model the structural relationships between foreground, background, and composite images, including alpha blending behavior and object-layer semantics. However, models trained solely on such data tend to produce unnatural or low-quality foreground layers. To overcome this, the second stage leverages high-quality unlayered image datasets to refine the composite generation branch. Through the shared backbone and coupled design, this refinement indirectly enhances the foreground generation branch, ultimately enabling the synthesis of high-quality layered images from limited supervision.

With extensive experiments including a user study, we demonstrate that \MethodName{} achieves high-quality layered image synthesis, where foreground and background layers are well harmonized, along with strong generalization compared to existing methods, as shown in \cref{fig:teaser}. 
\MethodName{} also enables diverse practical applications, such as foreground layer extraction from composite and background images, reference-based foreground generation, and multi-layered image synthesis.
Our main contributions can be summarized as:
\begin{itemize}
    \item We propose \MethodName{} that enables effective BG2FG layered image synthesis while addressing the dataset challenge.
    \item We introduce a dual-branch diffusion model that enables cross-modal knowledge transfer, and a semi-supervised learning strategy for improved foreground synthesis.
    \item Extensive experiments demonstrate that \MethodName{} outperforms existing methods in both visual quality and generalization, showing strong applicability to real-world scenarios. 
\end{itemize}

\section{Related Work}
\label{sec:relatedwork}

\paragraph{Layered Image Synthesis.}
Motivated by the practical value of layered image representations, interest in their automatic generation has grown rapidly. 
In this section, we review the architectural designs adopted by prior methods. 
Beyond the early GAN-based~\citep{furrygan} and CLIP-based~\citep{text2live} attempts, most recent methods leverage the generative power of pretrained diffusion models via fine-tuning.
Decomposition-based methods~\citep{layeringdiff,layerdecomp} adapt the input/output layers of diffusion models to decompose a composite image into foreground and background layers.

\begin{figure*}
    \centering
    \includegraphics[width=\linewidth]{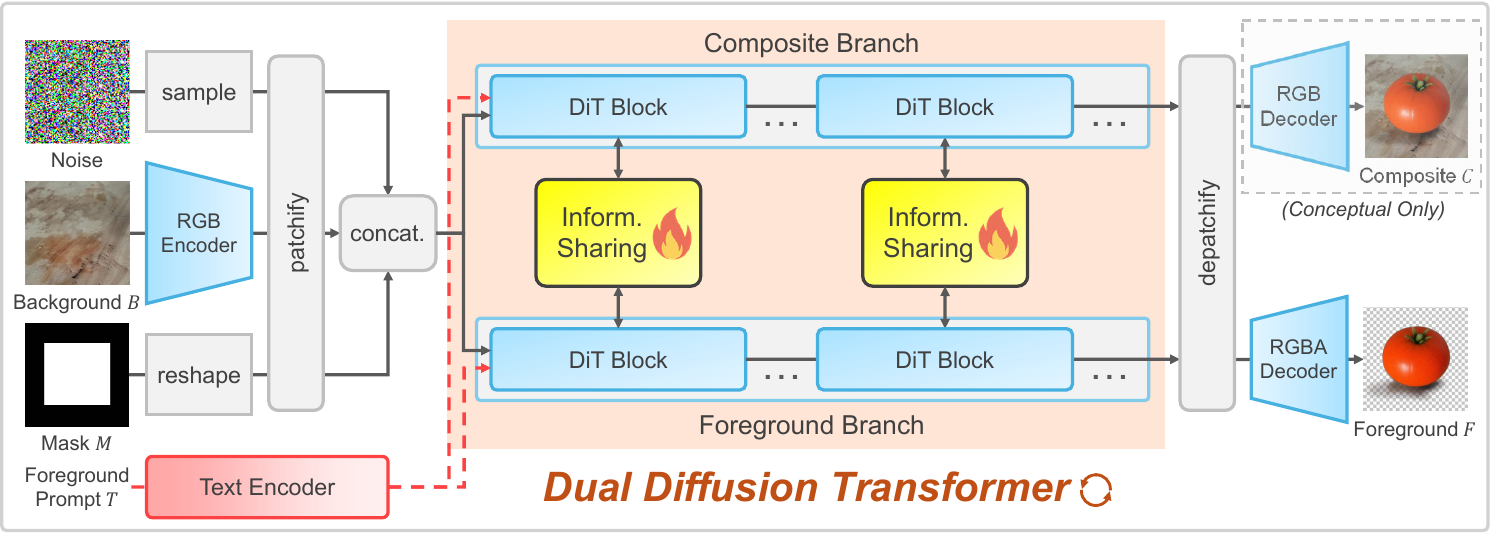}
    \caption{
    Overall framework of \MethodName{}.
    Given a background image $B$, a bounding box $M$, and a foreground prompt $T$, \MethodName{} generates a foreground RGBA layer $F$ through a bidirectionally interconnected dual-branch diffusion transformer. The composite branch produces a composite latent for knowledge transfer, but its output is not decoded. Only the foreground latent is decoded to obtain the final result.
    }
    \label{fig:pipeline}
\end{figure*}

Generation-based methods directly synthesize layers from random noise and text descriptions. 
Text2Layer~\citep{text2layer} introduces a unified latent space in which foreground and background are jointly embedded, enabling both layers to be generated in a single pass.
Subsequent methods~\citep{layerdiff,layerdiffuse,layerfusion,dreamlayer,psdiffusion} have established a dominant paradigm of multi-pass joint synthesis, allocating a dedicated generative pathway to each layer. 
For instance, LayerDiff~\citep{layerdiff} synthesizes layers from layer-specific text prompts while promoting both intra- and inter-layer interactions via a collaborative attention block, and LayerDiffuse~\citep{layerdiffuse} leverages layer-specific LoRA adapters for each layer and aggregates all attention vectors across all pathways to form a unified model.

More recent works~\citep{layerfusion,dreamlayer,psdiffusion} augment the design with an additional composite branch, leveraging cross-attention maps to inject global scene context into the individual layers.
Nevertheless, they remain tied to U-Net–based diffusion backbones and depend heavily on cross-attention maps.
In contrast, the only DiT-based work, ART~\citep{art}, adopts a fundamentally different strategy. It first generates a global layout, assigns latent tokens to spatial regions, and synthesizes all layers simultaneously with a diffusion transformer. While this design enables efficient region-wise generation, it inherently limits the ability to capture holistic visual effects spanning across multiple regions.

\paragraph{Object Insertion.}
Another closely related line of research is object insertion, which integrate an object into a target scene while ensuring object-level plausibility (pose, scale, and identity) and scene-level consistency (lighting, shadows, reflections, and other visual effects).
Recently, large-scale image generative models have motivated \emph{single framework solutions} that address all the subproblems simultaneously, directly producing composite images in a single synthesis. 
Among them, training-free methods~\citep{tf-icon,add-it} demonstrate that high-quality insertion results can be achieved by guiding the diffusion generation process.
A more dominant direction is to fine-tune diffusion models to insert a reference image into a given background~\citep{paintbyexample,objectstitch,controlcom,anydoor,imprint}. 
To improve realism, subsequent works curate interaction-aware data or exploit object-removal pairs \citep{dreamfuse,affordance-aware-insertion,insertanything,outsidebbox,objectdrop,erasedraw,ORIDa,omnipaint}. Recent extensions also explore multi-object insertion and multi-view identity cues \citep{multitwine,objectmate,dreamcom}, and layout-first then inpaint formulations \citep{smartmask,generative-location-modeling}.

However, most object insertion methods concentrate on composite image generation rather than learning disentangled representations.
As a result, the inserted objects cannot be easily separated, reused, or independently edited after synthesis.
In contrast, our approach explicitly generates a foreground RGBA layer which is an editable and reusable representation of the inserted object.

\section{Dual-Branch Diffusion Architecture}
\label{sec:method}

As introduced in \Sec{\ref{sec:introduction}}, we focus on the BG2FG problem in layered image synthesis by proposing \MethodName{}, which is illustrated in \Fig{\ref{fig:pipeline}}.
Given a background image $B$, a bounding-box mask $M$, and a foreground text description $T$, it generates a foreground RGBA layer $F=(F_{rgb}, F_\alpha)$ that contains the object specified by $T$ within $M$.
The composition follows the standard alpha blending equation:
\begin{equation}
C = F_{\alpha} \cdot F_{rgb} + (1-F_\alpha) \cdot B,
\end{equation}
where $F_{rgb}$ and $F_{\alpha}$ are the RGB color channels and corresponding opacity map of $F$, respectively, and $\cdot$ is element-wise multiplication.

To enable post-editing of the synthesized layered image, we define the foreground representation to satisfy the following properties:
(1) it includes not only the foreground object but also its associated visual effects, such as shadows and reflections, and
(2) it naturally harmonizes with the background when composited.
These properties allow lightweight and immediate editing in typical real-world scenes, where such effects remain largely consistent under moderate foreground transformations (e.g., repositioning or replacement).

To synthesize such foreground layers, 
\MethodName{} is to leverage the relatively easy-to-learn knowledge of composite image synthesis to guide the more challenging foreground synthesis, thereby alleviating the need for complex training data construction.
To achieve this, we adopt a dual-branch diffusion architecture that jointly generates a composite image and its foreground layer, with bidirectional information exchange ensuring semantic alignment between branches.

Specifically, we instantiate two parallel copies of a diffusion transformer and introduce an \emph{information-sharing} module based on a symmetric cross-attention layer. 
To detail the information-sharing module, let $H^{C}$ and $H^{F}$ be the intermediate features before the self-attention layer at each transformer block of the composite and foreground pathways, respectively.
We compute two cross-attentions by querying one pathway with keys and values from the other:

\begin{equation}
\small{
\begin{aligned}
Z^{C \to F}
= \mathrm{softmax}\!\left(\frac{Q^{F} (K^{C})^\top}{\sqrt{d}}\right) V^{C}, \quad
Z^{F \to C}
= \mathrm{softmax}\!\left(\frac{Q^{C} (K^{F})^\top}{\sqrt{d}}\right) V^{F},
\end{aligned}
}
\end{equation}
where $Q$, $K$, and $V$ denote the query, key, and value embeddings of each branch, while superscripts $F$ and $C$ indicate the foreground and composite pathways, respectively. $d$ is the embedding dimension.

For computing the cross-attention embeddings, we use shared and frozen base matrices $W_Q$, $W_K$, and $W_V$, initialized from the backbone’s self-attention weights, and introduce two \emph{direction-specific} LoRAs \(\{\Delta W_{\{\cdot\}}^{\,C \to F}, \Delta W_{\{\cdot\}}^{\,F \to C}\}\) to modulate these matrices. The resulting feature embeddings are computed as:
\begin{equation}
\begin{aligned}
Q^{C} &= H^{C}\!\left(W_Q + \Delta W_{Q}^{\,F \to C}\right),&
\hspace{-6pt}
Q^{F} &= H^{F}\!\left(W_Q + \Delta W_{Q}^{\,C \to F}\right), \\
K^{C} &= H^{C}\!\left(W_K + \Delta W_{K}^{\,C \to F}\right),&
\hspace{-6pt}
K^{F} &= H^{F}\!\left(W_K + \Delta W_{K}^{\,F \to C}\right), \\
V^{C} &= H^{C}\!\left(W_V + \Delta W_{V}^{\,C \to F}\right),&
\hspace{-6pt}
V^{F} &= H^{F}\!\left(W_V + \Delta W_{V}^{\,F \to C}\right).
\end{aligned}
\end{equation}

The two cross-attention messages are concatenated and passed through a lightweight MLP \(g(\cdot)\), whose outputs are split into two residuals $\Delta H^{C}$ and $\Delta H^{F}$ for each branch:
\begin{equation}
\begin{aligned}
[\Delta H^{C}, \Delta H^{F}] = g\bigl(\mathrm{concat}(Z^{F \to C}, Z^{C \to F})\bigr).
\end{aligned}
\end{equation}
We then integrate these residuals into each branch by adding them to the output of the self-attention layer of each branch:
\begin{equation}
\begin{aligned}
\widetilde{H}^{C} = \mathrm{SelfAttn}(H^{C}) + \Delta H^{C},\;
\widetilde{H}^{F} = \mathrm{SelfAttn}(H^{F}) + \Delta H^{F}.
\end{aligned}
\end{equation}

In our implementation, we adopt Flux-Fill~\citep{flux} as our backbone transformer, which is an image-inpainting diffusion transformer selected for its task affinity.
Keeping the backbone frozen, we train only the two \emph{direction-specific information-sharing} LoRAs (\(\{\Delta W_{\cdot}^{\,C \to F},\, \Delta W_{\cdot}^{\,F \to C}\}\)) and a lightweight MLP \(g(\cdot)\).
To encode background and composite inputs into the latent space, we use the pretrained RGB VAE of Flux-Fill, while the RGBA foreground layer is encoded and decoded using an RGBA VAE fine-tuned on our dataset.
Additional details on RGBA VAE fine-tuning and other implementation details including LoRA configurations are provided in the supplementary material.

At inference time, the diffusion denoising process operates in the latent space of the pretrained VAE.
A random noise tensor is first channel-wise concatenated with the conditioning inputs (the background latent and a bounding box mask reshaped following Flux-Fill).
The concatenated tensor is then duplicated along the batch dimension and passed through the diffusion transformer with text embeddings computed from foreground descriptions. 
After iterative denoising, the model produces a composite latent and a foreground latent.
Finally, only the foreground latent is decoded into the image domain using the RGBA VAE.

\begin{figure*}
    \centering
    \includegraphics[width=\linewidth]{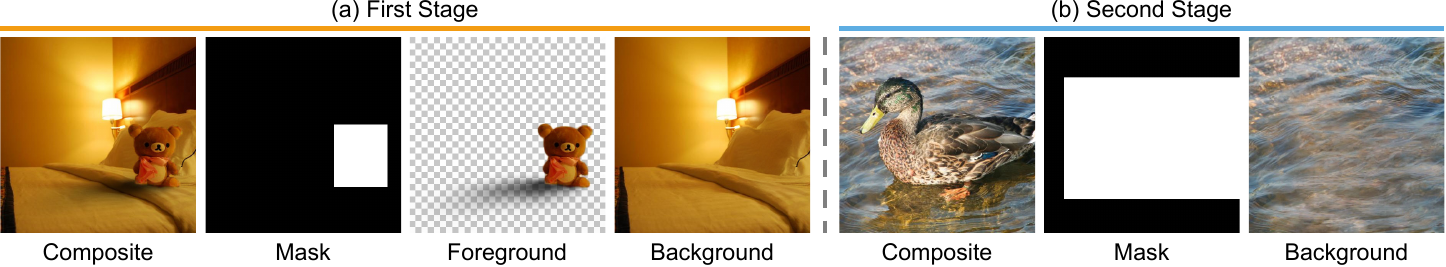}
    \caption{Sample images from the training corpora used in each stage.}
    \label{fig:dataset}
\end{figure*}

\section{Two-stage Semi-Supervised Learning}

To train \MethodName{}, the most straightforward dataset would comprise foreground, background, composite images, and bounding-box masks.
Yet, as discussed in \Sec{\ref{sec:introduction}}, constructing a layered image dataset in practice is highly challenging.
To overcome this challenge, we propose a two-stage semi-supervised learning strategy that avoids explicit reliance on ground-truth foreground layers.
In brief, we first train \MethodName{} on a simulation-based synthetic dataset that provides all modalities, allowing the model to learn their structural relationships.
We then fine-tune it on a high-quality unlayered image dataset to enhance the composite synthesis, which in turn improves the foreground synthesis through the information-sharing module. In the following sections, we provide the details of the datasets and the training losses used in each stage.

\subsection{Representation Learning with Compositional Consistency}
\MethodName{} first learns modality-specific representations and their compositional relationships using a synthetic dataset composed of foreground, background, and composite images, along with bounding-box masks.
To construct the dataset, we adopt an approach similar to LayerDecomp~\citep{layerdecomp} on the RORem object removal dataset~\citep{rorem}, which provides composite images, object masks, and background images where both objects and their visual effects have been removed. 

For each composite image $C$, we first extract a foreground layer $F$ from the object mask $M$ using a matting technique~\citep{vitmatte}, then augment it to include simulated shadows using an internal shadow generation network, which takes the foreground layer placed on a white background as input and renders shadow effects into the image.
The augmented foreground layer $\hat{F}$ is recomposited with the background layer $B$ to form a new composite image $\hat{C}$.
Finally, we replace the original object mask with a bounding-box mask $\hat{M}$, resulting in a training corpus of $\{{\hat{F}, B, \hat{M}, \hat{C}}\}$, as shown in \cref{fig:dataset} (a).
We provide the details of the shadow generation network and visual examples of the dataset construction pipeline in the supplementary material.
For notational simplicity, we drop the hat hereafter. 

Our data construction pipeline produces image layers with precise pixel-level alignment and embeds simulated visual effect into the foreground layer, even though these effects are less realistic and not harmonized with the background layer. Unlike LayerDecomp~\citep{layerdecomp}, which constructs composites by randomly pairing foregrounds with backgrounds, our method derives foregrounds directly from real composite images, thereby offering more reliable supervision for object placement and scale.

Using the synthetic dataset, we employ a flow-matching loss.
Let $z_{F}$, $z_B$, and $z_{C}$ denote the latent representations of the foreground, background, and composite images, respectively. 
At each training iteration, we sample Gaussian noise $z_0 \sim \mathcal{N}(0,I)$ and linearly interpolate it with each data latent $z_1 \in \{z_{F}, {z_{C}}\}$ at a random time step $t\sim\mathcal{U}(0,1)$: $z^{(t)} = (1-t) z_0 + t z_1$
The flow-matching loss is defined as:
\begin{equation}
\begin{aligned}
\mathcal{L}_{\mathrm{flow}}
= \mathbb{E}_{t,z_0,z_1}\!\Big[
\big\|v^{\ast}_{C} &- v_{\theta}^{C}(z_{C}^{(t)},z_{F}^{(t)}, t, \kappa)\big\|_2^2 
\\+
\big\|v^{\ast}_{F} &- v_{\theta}^{F}(z_{C}^{(t)},z_{F}^{(t)}, t, \kappa)\big\|_2^2
\Big],
\end{aligned}
\label{eq:flow}
\end{equation}
where $v^{\ast}_{C}$ and $v^{\ast}_{F}$ are the ground-truth velocity fields (given by $v^{\ast} = z_1 - z_0$). $v_{\theta}^{C}$ and $v_{\theta}^{F}$ are velocity fields predicted by the composite and foreground branches, respectively.
Note that $v_{\theta}^{C}$ and $v_{\theta}^{F}$ are computed from both $z_{C}^{(t)}$ and $z_{F}^{(t)}$, as the composite and foreground branches are connected through information-sharing modules.
Here, $\kappa$ represents the conditioning inputs, including the foreground text description $T$, the background latent $z_B$, and a reshaped mask $M$.

To further enforce compositional consistency between the branches, we introduce a composition loss, which is defined as:
\begin{equation}
\mathcal{L}_{\mathrm{comp}} = \big\|\Phi(\hat{z}_{F},z_B) - \hat{z}_{C}\big\|_1,
\label{eq:L_comp}
\end{equation}
where $\hat{z}_{C}$ and $\hat{z}_{F}$ are the model's one-step denoised estimates at time $t$, (i.e., the predicted clean latents from $z_C^{(t)}$ and $z_F^{(t)}$). The function $\Phi$ composites the foreground latent $\hat{z}_{F}$ and the background latent $z_B$ in the latent space.
We implement $\Phi$ as a neural network, pre-trained separately and then fixed during the training of our framework. More details of $\Phi$ is provided in the supplementary material.
Finally, the overall objective of the first stage is defined as:
\begin{equation}
\mathcal{L}_{\mathrm{first}} = \mathcal{L}_{\mathrm{flow}} + 0.1  \mathcal{L}_{\mathrm{comp}}.
\end{equation}

\subsection{Realism Enhancement with Distribution Regularization}
In the second stage, we enhance the realism of the generated images using a dataset of natural, unlayered images that capture diverse real-world effects.
Specifically, we construct high-quality paired background and composite images and use them to supervise only the composite branch. The resulting improvements then propagate to the foreground branch through the information-sharing modules, aligning the foreground with the refined composite and calibrating its visual effects.

To be specific for the dataset construction, we remove foreground objects from composite images of the RORem dataset~\citep{rorem} to generate corresponding background images as shown in \cref{fig:dataset} (b).
Although the RORem dataset already provides background images obtained by inpainting the masked regions, these often still contain residual visual effects of the foreground objects. To obtain cleaner backgrounds, we instead employ ObjectClear~\citep{objectclear}, a recent object-removal method that removes both objects and their associated effects. 
This construction pipeline is scalable, which makes it easy to further expand the training corpus.

Based on the dataset, we define a realism-boosting loss that supervises only the composite branch:
\begin{equation}
\mathcal{L}_{\mathrm{real}}
= \mathbb{E}_{t,z_0,z_1}\left[
          \big\|v^{\ast}_{C} - v_{\theta}^{C}(z_C^{(t)}, z_{S}^{(t)}, t, \kappa)\big\|_2^2 
\right],
\end{equation}
where $v^{\ast} = z_1 - z_0$.
As no ground-truth foreground layer $F$ is available, we instead use a \emph{surrogate} input $S$ to compute the surrogate latent $z_{S}^{(t)}$ for the input of the foreground branch, where $S$ is obtained by applying an image matting network~\citep{vitmatte} to the composite image $C$ based on the mask $M$.
We empirically find that this surrogate input has negligible influence on learning dynamics.

While the composite supervision improves realism, relying on it alone may cause the foreground distribution to drift away from the distribution established in the previous stage, degrading the quality of the foreground layer.
To mitigate this, we introduce additional supervision for the foreground branch using the synthetic dataset of the previous stage. 
Although the dataset is not fully realistic, it benefits from the pixel-wise alignment across $\{F, B, C\}$.
Using this dataset, we introduce a foreground regularization loss defined with only the foreground branch term of the flow-matching objective in \cref{eq:flow}:
\begin{equation}
\mathcal{L}_{\mathrm{reg}}
= \mathbb{E}_{t,z_0,z_1}\left[
          \big\|v^{\ast}_{F} - v_{\theta}^{F}(z_C^{(t)},z_F^{(t)}, t, \kappa)\big\|_2^2
\right].
\end{equation}
During training iterations, we alternate between optimizing the realism-boosting loss and the foreground regularization loss, each combined with the composition loss. Specifically, we optimize $\mathcal{L}_{\mathrm{real}} + 0.5\mathcal{L}_\mathrm{comp}$ with an 80\% probability and $\mathcal{L}_{\mathrm{reg}} + 0.5\mathcal{L}_\mathrm{comp}$ with a 20\% probability at each iteration.

\section{Experiments}
\label{sec:experiments}

\subsection{Layered Image Synthesis Evaluation}
\paragraph{Baselines.}
We compare the quality of the layered images generated by \MethodName{} and existing approaches on multiple datasets, including SAM-FB~\citep{affordance-aware-insertion}, RORem~\citep{rorem}, and AnyInsertion~\citep{insertanything}. 
Each dataset provides background images and masks specifying the regions where foreground objects should be placed. For consistency, all masks are converted into bounding-box form.
All input images (backgrounds and masks) are resized such that the shorter side is $512$ pixels before being fed into the models. For the target foreground text descriptions, we use the captions provided by the SAM-FB and AnyInsertion datasets, while those for the RORem dataset are generated using BLIP2~\citep{blip2}.

Since decomposition-based methods are not designed for BG2FG synthesis, we instead evaluate their outputs by applying the extracted layers to the ground-truth composite images.
Specifically, we first compute foreground masks using SAM2~\citep{sam2} from the bounding-box masks, and estimate foreground layers using LayeringDiff~\citep{layeringdiff}.
We then estimate the background layers using ObjectClear~\citep{objectclear}, since LayeringDiff often leaves residual visual effects in the background. 
We refer to this combination as LD+OC.
For LayeringDiff and ObjectClear, we use the code provided by the authors.
We do not include LayerDecomp~\citep{layerdecomp} in our comparison as its source code is not publicly available. However, the quality of its background estimates are expected to be comparable to ObjectClear, given the close similarity in their training strategies.

For generation-based methods, we only compare with LayerDiffuse~\citep{layerdiffuse}, as it is the only method with publicly available code. Specifically, we adopt its best-performing SDXL-based two-stage variant, which first generates a composite image and then estimates the foreground layer conditioned on the background and the composite.
For other generation-based models, we provide a discussion based on their reported results in the supplementary material.

\newlength{\origtabcolsep}
\setlength{\origtabcolsep}{\tabcolsep}
\setlength{\tabcolsep}{2pt} 

\begin{table}[t]


\caption{Quantitative evaluation of existing layered image synthesis methods on RORem dataset~\citep{rorem}.
$^\dagger$Note that KID may yield small negative values due to sampling variance. These should be interpreted as values close to zero.
}
\label{tbl:exp_quantitative}
\scalebox{0.9}{
\begin{tabular}{c|c|ccccc}
\hline

 & \multicolumn{1}{c|}{FG} & \multicolumn{5}{c}{Composite (or Background for LD+OC)} \\
\hline
Method                                & CLIP$\uparrow$
                                      & MUSIQ$\uparrow$ 
                                      & MANIQA$\uparrow$
                                      & KID$_{\times1000}$$\downarrow$ 
                                      & DINO$\uparrow$
                                      & HPSv3$\uparrow$\\
\hline
LD+OC                                 & 0.701
                                      & 63.485
                                      & 0.382
                                      & \textbf{-0.2}$^\dagger$
                                      & \textbf{0.961} 
                                      & 2.91\\
LayerDiffuse                          & \textbf{0.770}
                                      & 67.221
                                      & 0.425
                                      & 11.0
                                      & 0.467 
                                      & -3.03\\
Ours                                  & 0.721
                                      & \textbf{67.886}
                                      & \textbf{0.437}
                                      & 0.2
                                      & 0.875 
                                      & \textbf{3.70}\\
\hline

\end{tabular}}
\end{table}
\setlength{\tabcolsep}{\origtabcolsep}

\begin{table}[t]
\centering
\caption{
Results of the user study on 20 examples, averaged over 18 participants. Higher scores indicate better quality (1–5).}
\label{tbl:exp_user_study}
\scalebox{0.9}{
\begin{tabular}{lccc}
\toprule
\textbf{Metric} & LD+OC & LayerDiffuse & \MethodName{} \\ \midrule
(1) FG--Text Alignment         & \textbf{4.02}    &  2.96  &   3.99\\
(2) FG--Visual Effects         & 3.49    &  2.33  &   \textbf{3.90}\\
(3) FG--Image Quality          & 3.61    &  2.29  &   \textbf{3.88}\\
(4) Composite--Placement       & 4.11    &  1.82  &   \textbf{4.32}\\
(5) Composite--Harmonization   & 3.04    &  1.28  &   \textbf{3.68}\\
(6) Composite--Image Quality   & 3.33    &  1.64  &   \textbf{3.74}\\ \midrule
\textbf{Average}               & 3.60    &  2.05  &   \textbf{3.92}\\
\bottomrule
\end{tabular}
}
\end{table}

\paragraph{Qualitative Comparison.}
\Fig{\ref{fig:exp_qualitative}} presents a qualitative comparison. 
LD+OC fails to capture fine foreground details (e.g., under the train body), and the recovered background does not faithfully restore the rail structures.
The foreground objects generated by LayerDiffuse are oversized and inconsistent with the background context.
This indicates that the synthetic dataset used for its training does not generalize well to real-world backgrounds, leaving the model uncertain about how to position objects, and simply producing large, centrally placed ones that follow the typical distribution of RGBA images.
In contrast, our method generates foreground objects that are contextually appropriate, visually consistent, and equipped with proper visual effects, yielding more coherent composites.

\paragraph{Quantitative Comparison.}
Quantitative evaluation of generated layered images is inherently difficult, as ground-truth layered images are not available. In this work, we evaluate both the generated foreground layer and the composite image obtained by combining the generated foreground with the input background.
For decomposition-based methods, whose input is a composite image, we evaluate their estimated background layers instead.
For foreground evaluation, we use CLIP score~\citep{clipscore} to measure alignment between the generated (or decomposed) foreground and the input foreground caption. 
For composite (or background) evaluation, we employ non-reference aesthetic quality metrics (MUSIQ~\citep{omnipaint} and MANIQA~\citep{omnipaint}), as well as KID~\citep{kid} to measure distributional similarity against the target distribution. We further report the DINO score~\citep{dinov2} to directly compare with the ground-truth. 
Finally, to quantify human preference, we adopt the HPSv3~\cite{hpsv3}, which is designed to align with human judgments.
Since HPSv3 requires a text description as input, we use captions generated by BLIP2~\cite{blip2} on ground-truth.

We report results on 2,000 images from the RORem dataset in \Tbl{\ref{tbl:exp_quantitative}}, while results on other datasets are provided in the supplementary material.
LayerDiffuse~\citep{layerdiffuse} attains the highest CLIP scores, as it tends to generate large foreground objects that dominate the image and are thus favored by the metric. However, its composite evaluation scores are substantially worse than other methods, especially in KID and DINO, indicating that the generated composites deviate significantly from the ground-truth distribution.
LD+OC~\citep{layeringdiff,objectclear} achieves strong performance in both KID and DINO, highlighting its strength in background estimation.
Our method achieves a higher CLIP score than LD+OC, while also outperforming LayerDiffuse by a large margin in composite evaluation.
Moreover, our method achieves the highest HPSv3 score among all methods.

\paragraph{User Study.}
We designed a human preference study recruiting 18 volunteers from our institution and 20 test cases. In each test case, participants were presented with the foreground prompt along with the generated foreground, background, and composite images produced by LD+OC~\cite{layeringdiff,objectclear}, LayerDiffuse~\citep{layerdiffuse}, and \MethodName{}.
Participants rated six aspects: (1) alignment between the foreground and text description, (2) correctness of visual effects (e.g., shadows and reflections) in the foreground, (3) visual quality of the foreground, (4) appropriateness of object placement in the composite, (5) harmonization between the foreground and background, and (6) overall visual quality of the composite. 
As summarized in \Tbl{\ref{tbl:exp_user_study}}, \MethodName{} achieves consistently high ratings across all criteria, demonstrating a clear preference among participants.
The detailed questionnaire is provided in the supplementary material.

\subsection{Comparison with Object Insertion Methods}
\MethodName{} is conceptually related to naturally adding new objects into a given scene. To evaluate this capability, we compare it against two recent models designed for similar purposes: the image inpainting model Flux-Fill~\citep{flux} and the object insertion method PaintByInpaint~\citep{paintbyinpaint}.
For Flux-Fill, we adopt the default setting. 
Since PaintByInpaint does not take an input mask and uses an instruction-based text description, we prepend the word “add” to the foreground caption before providing it to the model. 
\setlength{\tabcolsep}{2pt} 

\begin{table}[t]
\caption{
Quantitative comparison with an object insertion method PaintByInpaint~\cite{paintbyinpaint} and an image inpainting method Flux-Fill~\cite{flux} on RORem datset~\citep{rorem}.}
\label{tbl:exp_quantitative_inpainting}
\centering
\scalebox{0.9}{

\begin{tabular}{c|ccccc}
\hline
Method                                & MUSIQ$\uparrow$
                                      & MANIQA$\uparrow$
                                      & KID$_{\times1000}$$\downarrow$
                                      & DINO$\uparrow$  
                                      & HPSv3$\uparrow$\\
\hline
PaintByInpaint & 65.516
                                      & 0.358
                                      & 0.889
                                      & 0.824 
                                      & 1.65\\
Flux-Fill                             & 66.132
                                      & 0.425
                                      & 0.934
                                      & 0.857
                                      & 2.74\\
Ours                                  & \textbf{67.886}
                                      & \textbf{0.437}
                                      & \textbf{0.2}
                                      & \textbf{0.875} 
                                      & \textbf{3.70}\\
\hline
\end{tabular}}
\end{table}
\setlength{\tabcolsep}{\origtabcolsep}

\begin{figure*}[t]
    \centering
    \includegraphics[width=\linewidth]{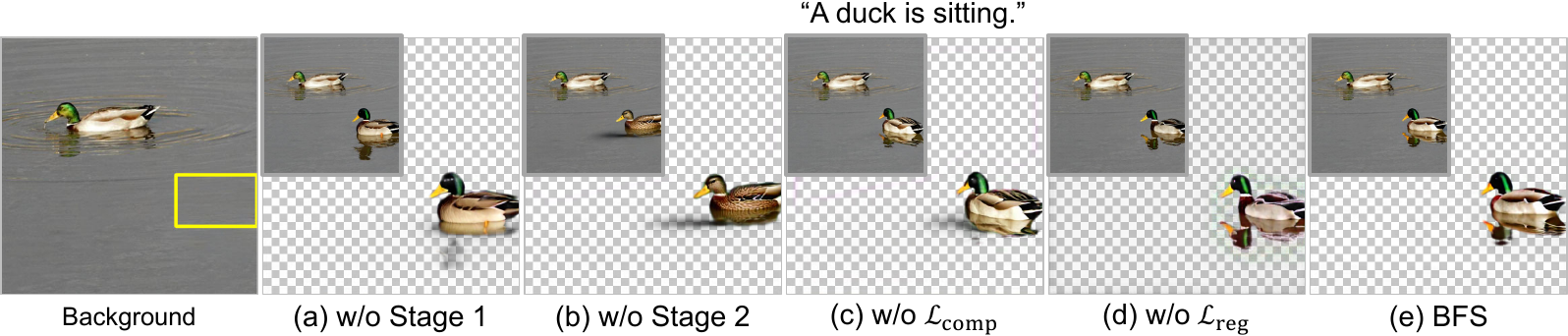}
    \caption{Ablation study. Qualitative comparison of \MethodName{} with different components removed. The yellow box denotes the input mask and the inset in the top-left shows the output of the composite branch for reference.}
    \label{fig:exp_ablation}
\end{figure*}

\begin{figure*}[t]
\begin{minipage}{\linewidth}
    \centering
    \vspace{0.3cm}
    \includegraphics[width=1\linewidth]{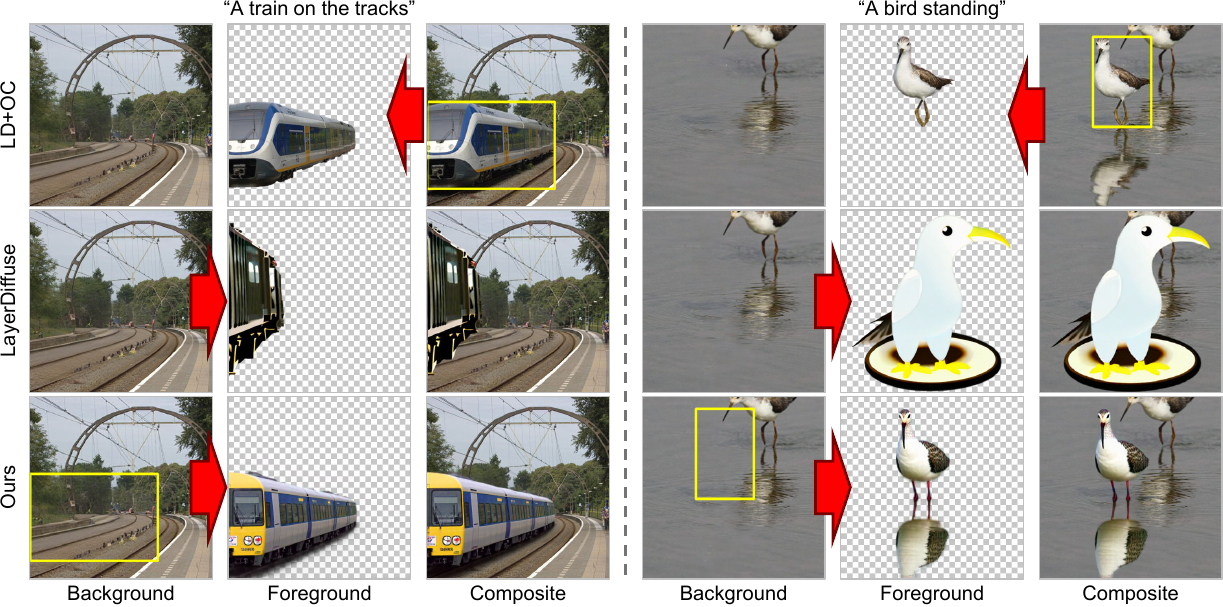}
    \caption{
    Qualitative comparison with existing layered image synthesis approaches, LD+OC~\citep{layeringdiff,objectclear} and LayerDiffuse~\citep{layerdiffuse}. 
    The yellow box denotes the input mask, which is not used by LayerDiffuse.
    }
    \label{fig:exp_qualitative}
    \vspace{0.2cm}
\end{minipage}

%
\begin{minipage}{\linewidth}
    \centering
    \vspace{0.2cm}
    \includegraphics[width=1\linewidth]{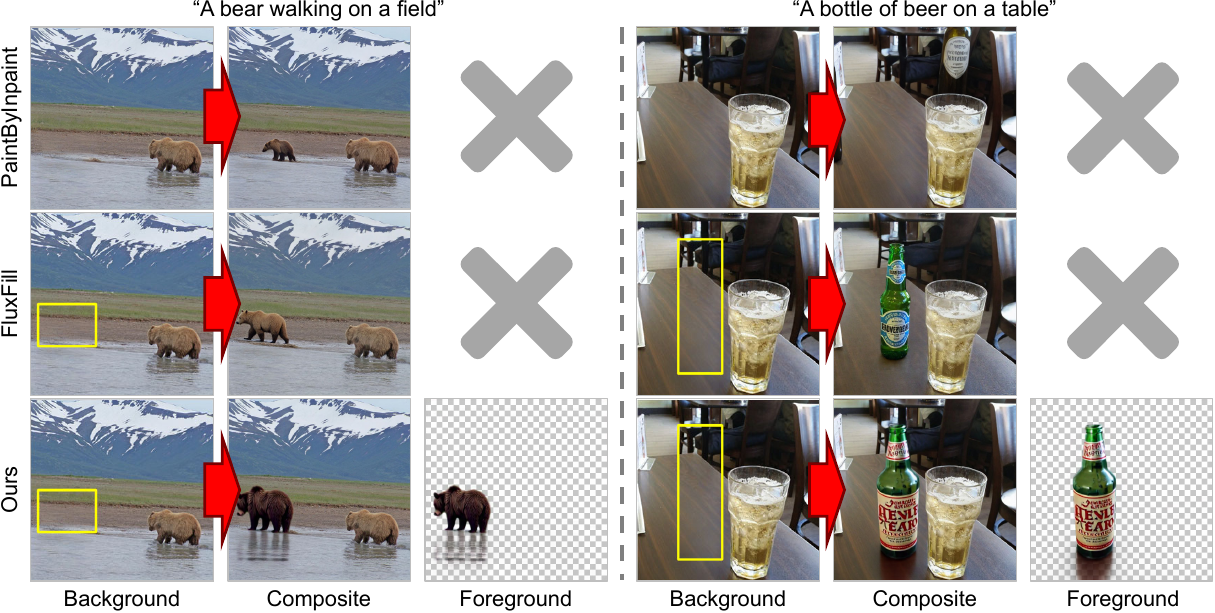}
    \caption{Qualitative comparison with an object insertion method, PaintByInpaint~\citep{paintbyinpaint}, and an image inpainting method, Flux-Fill~\citep{flux}. In contrast to these methods, ours additionally produces explicit and reusable foreground layers. The yellow box denotes the input mask.}
    \label{fig:exp_qual_inpainting}
    \vspace{0.3cm}
\end{minipage}
\end{figure*}

\Fig{\ref{fig:exp_qual_inpainting}} presents a qualitative comparison of the composites generated by these baselines and our approach. For reference, we also show the foreground layers produced by \MethodName{}. PaintByInpaint often struggles to place objects at semantically suitable locations, while FluxFill generates realistic images but performs editing strictly within the input bounding box, often resulting in composites that lack proper visual effects. In contrast, our method synthesizes high-quality composite images, while additionally producing explicit and reusable foreground layers.


\Tbl{\ref{tbl:exp_quantitative_inpainting}} summarizes the quantitative evaluation of the generated composite images on the RORem dataset~\citep{rorem}. Our method achieves the best scores across all metrics, demonstrating that \MethodName{} produces natural and coherent composites that are on par with state-of-the-art insertion models trained solely on real-world supervision.


\begin{figure}[t]
    \centering
    \includegraphics[width=\linewidth]{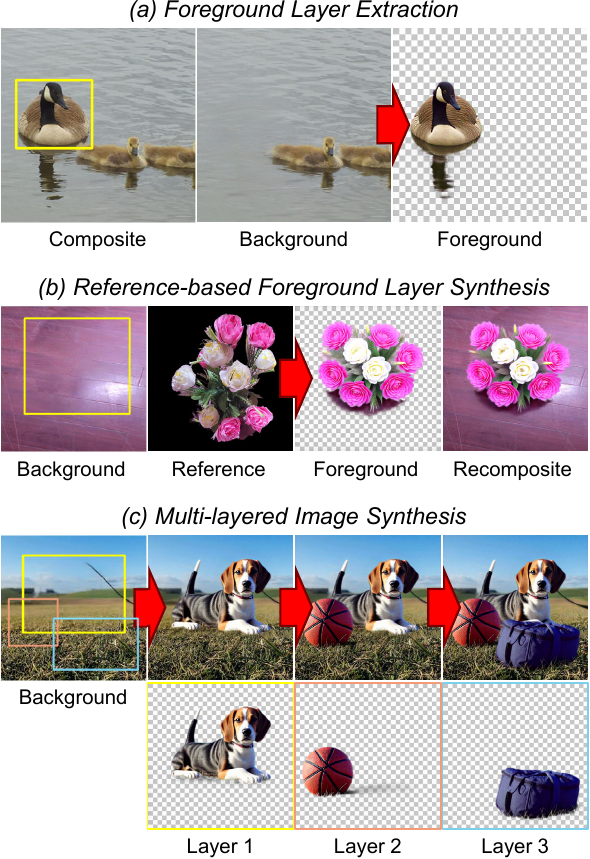}
    \caption{(a) Foreground layer extraction from a composite image given its corresponding background image with the target foreground removed. (b) Reference-based foreground layer synthesis guided by a reference image. (c) Multi-layered image synthesis achieved through the sequential application.
    }
    \label{fig:exp_application}
\end{figure}

\subsection{Ablation Study}
We conduct an ablation study to assess the contribution of each component in \MethodName{}.
\Fig{\ref{fig:exp_ablation}} presents qualitative comparisons of different model variants given the same background and mask (highlighted in yellow), with the outputs of the composite branch in the inset.
Without Stage~1 representation learning, directly training Stage~2 degrades foreground quality, yielding inferior reflection quality. 
Omitting Stage~2 generates composites with unnatural shadows that propagate to the foreground layer.
Removing the composition loss breaks consistency between the composite and foreground, causing unnatural reflections. Excluding the regularization step in Stage~2 collapses the RGBA distribution, yielding green halo regions around the foreground.
In contrast, our full model successfully aligns both branches and produces high-quality RGBA layers.

\subsection{Further Applications}

\MethodName{} can be extended to other applications as shown in \Fig{\ref{fig:exp_application}}. 
First, it enables \emph{foreground extraction} from existing images by feeding a noise-perturbed composite into the composite branch. Unlike conventional matting, this produces a foreground layer that preserves both the object and its visual effects (a). 
Second, it supports \emph{reference-based foreground synthesis} by replacing the foreground text embedding with that of a reference image via a Flux-Redux adapter, allowing the synthesized foreground to inherit the reference’s characteristics (b). 
Lastly, \MethodName{} extends to \emph{multi-layered image synthesis}, generating scenes with multiple foreground layers (c). 
More results appear in the supplementary material.

\subsection{Post-editing of Synthesized Layered Images}
\begin{figure*}
    \centering
    \includegraphics[width=0.855\linewidth]{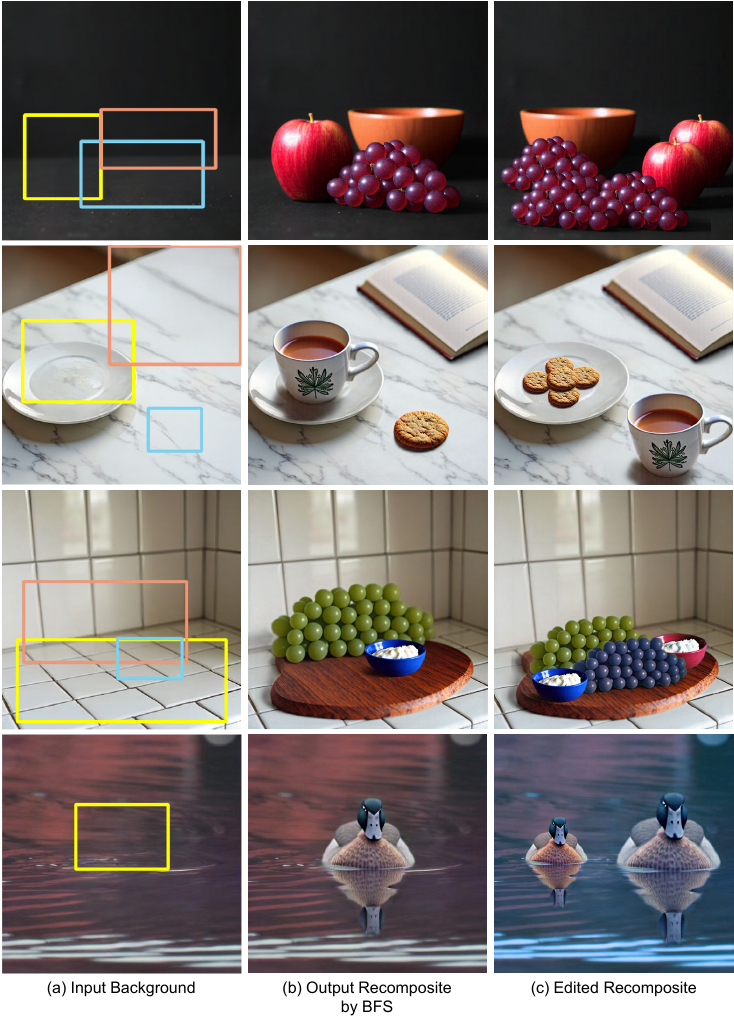}
    \caption{Post-hoc layered image editing. Input background images (a) are first extended into richer scenes by adding diverse objects with \MethodName{} (b), and are then further manipulated through layer‑wise edits (c).}
    \label{fig:exp_editing}
\end{figure*}

The layered images synthesized by \MethodName{} naturally support flexible post-hoc editing. 
\Fig{\ref{fig:exp_editing}} illustrates several examples where input background images (a) are first extended into a richer scene by adding multiple layers with \MethodName{} (b), and then further manipulated by editing each individual layer (c). All edits are performed with standard tools in Adobe Photoshop, demonstrating that the synthesized layered representations integrate seamlessly into conventional raster-graphics workflows and enable practical image generation.

\section{Conclusion}

We present \MethodName{}, a novel generation-based framework for high-quality layered image synthesis.
To overcome data scarcity, we transfer the relatively easy-to-learn knowledge of composite synthesis to foreground synthesis through a dual-branch framework that enables bidirectional knowledge exchange between the two branches.
We also propose a two-stage training scheme that utilizes a high-quality unlayered image dataset to further enhance foreground quality.
Extensive experiments demonstrate that \MethodName{} produce high-quality layered images, outperforming prior methods.

\paragraph{Limitations.}
\MethodName{} has certain limitations. 
Its dual-branch design increases inference time, as generating a single output with BFS takes 25 seconds compared to 12 seconds with Flux-Fill backbone.
In addition, our training data synthesis pipeline relies on an object-centric matting formulation over general images, which introduces two limitations. First, the relative scarcity of high-quality alpha mattes for transparent objects makes transparent object insertion challenging. We provide further discussion and examples in the supplementary material. Second, this formulation struggles to represent spatially diffuse or non-object-centric phenomena, making scene-wide effects such as fog or raindrops difficult to generate.
Finally, although \MethodName{} can synthesize plausible visual effects, these effects are not guaranteed to be physically accurate. In future work, we plan to incorporate rendering toolchains that can model more complex and physically grounded effects.

\begin{acks}
This work was supported by the Institute of Information \& Communications Technology Planning \& Evaluation (IITP) grants funded by the Korea government (MSIT) (No. RS-2019-II191906, Artificial Intelligence Graduate School Program (POSTECH); No. RS-2024-00395401, Development of VFX Creation and Combination Using Generative AI; No. RS-2024-00457882, AI Research Hub Project).
\end{acks}

\clearpage
\bibliographystyle{ACM-Reference-Format}
\bibliography{main}

\end{document}